\documentclass[10pt,twocolumn,letterpaper]{article}

\usepackage{cvpr}
\usepackage{times}
\usepackage{epsfig}
\usepackage{graphicx}
\usepackage{amsmath}
\usepackage{amssymb}

\usepackage{multirow}
\usepackage{stfloats}
\usepackage{tabularx}
\usepackage{algorithm}
\usepackage{algpseudocode}
\usepackage{subcaption}
\usepackage[usenames, dvipsnames]{color}

\definecolor{whitesmoke}{rgb}{0.96, 0.96, 0.96}
\definecolor{antiquewhite}{rgb}{0.98, 0.92, 0.84}

\DeclareMathOperator*{\argmin}{arg\,min}

\usepackage[pagebackref=true,breaklinks=true,letterpaper=true,colorlinks,bookmarks=false]{hyperref}

\cvprfinalcopy % *** Uncomment this line for the final submission

\def\cvprPaperID{1525} % *** Enter the CVPR Paper ID here

\begin{document}
%%%%%%%%%%%%%%%%my commands%%%%%%%%%%%%%%%%%%%%%%%%%%%
\newcommand{\figthree}[3]{
  \subfigure{  
  \begin{minipage}{5cm}
    \centering   
    \includegraphics[scale=0.4]{#1}            
  \end{minipage}
  }
   \quad \quad
  \subfigure{  
    \begin{minipage}{5cm}
      \centering    
      \includegraphics[scale=0.4]{#2}                
     \end{minipage}
   }
   \quad \quad
   \subfigure{
    \begin{minipage}{5cm}
     \centering    \includegraphics[scale=0.4]{#3}    
    \end{minipage}
   }
}
\newcommand{\figtwo}[2]{
  \subfigure{  
  \begin{minipage}{8cm}
    \centering   
    \includegraphics[scale=0.6]{#1}            
  \end{minipage}
  }
   \quad \quad
  \subfigure{  
    \begin{minipage}{8cm}
      \centering    
      \includegraphics[scale=0.6]{#2}                
     \end{minipage}
   }
}

\newcommand{\softnms}{soft-NMS~\cite{softnms}}
\newcommand{\ournms}{var voting}
\newcommand{\Ournms}{Variance Voting}
\newcommand{\ourloss}{KL Loss}
\newcommand{\datasetvar}{ambiguity}
\newcommand{\datasetvars}{ambiguities}
\newcommand{\Datasetvars}{Ambiguities}
\newcommand{\modelvar}{uncertainty}
\newcommand{\modelvars}{uncertainties}

\title{Bounding Box Regression with Uncertainty for Accurate Object Detection}

% Authors at the same institution
\author{Yihui He$^1$ \hspace{0.5cm} Chenchen Zhu$^1$ \hspace{0.5cm} Jianren Wang$^1$ \hspace{0.5cm} Marios Savvides$^1$ \hspace{0.5cm} Xiangyu Zhang$^2$\\
$^1$Carnegie Mellon University \hspace{2cm} $^2$Megvii Inc. (Face++)\\
{\tt\small \{he2,chenchez,jianrenw,marioss\}@andrew.cmu.edu \hspace{0.5cm} zhangxiangyu@megvii.com \hspace{0.5cm}}
}
% Authors at different institutions
% \author{First Author \\
% \and
% Second Author \\
\twocolumn[{%
\maketitle
% \ifcvprfinal\thispagestyle{empty}\fi
\begin{center}
    \includegraphics[width=\linewidth]{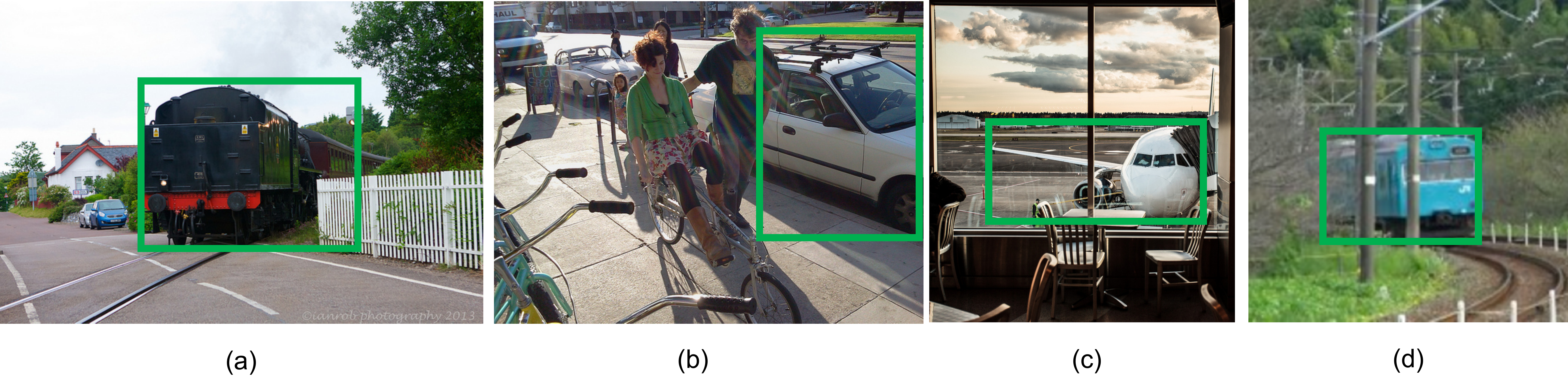}\\
  \captionof{figure}{In object detection datasets, the ground-truth bounding boxes have inherent ambiguities in some cases. The bounding box regressor is expected to get smaller loss from ambiguous bounding boxes with our \ourloss. (a)(c) The \datasetvars\ introduced by inaccurate labeling. (b) The \datasetvars\ introduced by occlusion. (d) The object boundary itself is ambiguous. It is unclear where the left boundary of the train is because the tree partially occludes it. 
    (\textit{better viewed in color})}
\label{fig:dataset}
\end{center}
}]

\begin{abstract}
Large-scale object detection datasets (\eg, MS-COCO) try to define the ground truth bounding boxes as clear as possible. However, we observe that \datasetvars\ are still introduced when labeling the bounding boxes. In this paper, we propose a novel bounding box regression loss for learning bounding box transformation and localization variance together. Our loss greatly improves the localization accuracies of various architectures with nearly no additional computation. The learned localization variance allows us to merge neighboring bounding boxes during non-maximum suppression (NMS), which further improves the localization performance. On MS-COCO, we boost the Average Precision (AP) of VGG-16 Faster R-CNN from 23.6\% to \textbf{29.1\%}. More importantly, for ResNet-50-FPN Mask R-CNN, our method improves the AP and AP$^{90}$ by \textbf{1.8\%} and \textbf{6.2\%} respectively, which significantly outperforms previous state-of-the-art bounding box refinement methods. 
% Our code and models are available at \href{https://github.com/yihui-he/KL-Loss}{github.com/yihui-he/KL-Loss}
Our code and models are available at github.com/yihui-he/KL-Loss

\end{abstract}

\section{Introduction}%%%%%%%%%%%%%%%%%%%%%%%%%%%%%%%%%%%%%%%%%
Large scale object detection datasets like ImageNet~\cite{imagenet}, MS-COCO~\cite{coco} and CrowdHuman~\cite{crowdhuman} try to define the ground truth bounding boxes as clear as possible.

However, we observe that the ground-truth bounding boxes are inherently ambiguous in some cases. The \datasetvars\ makes it hard to label and hard to learn the bounding box regression function.
Some inaccurately labeled bounding boxes from MS-COCO are shown in Figure~\ref{fig:dataset} (a)(c). When the object is partially occluded, the bounding box boundaries are even more unclear, shown in Figure~\ref{fig:dataset} (d) from YouTube-BoundingBoxes~\cite{youtube}.

Object detection is a multi-task learning problem consisting of object localization and object classification. Current state-of-the-art object detectors (\eg, Faster R-CNN~\cite{faster}, Cascade R-CNN~\cite{cascade} and Mask R-CNN~\cite{mask}) rely on bounding box regression to localize objects. 
 
\begin{figure}[t]
    \begin{center}
    \includegraphics[width=\linewidth]{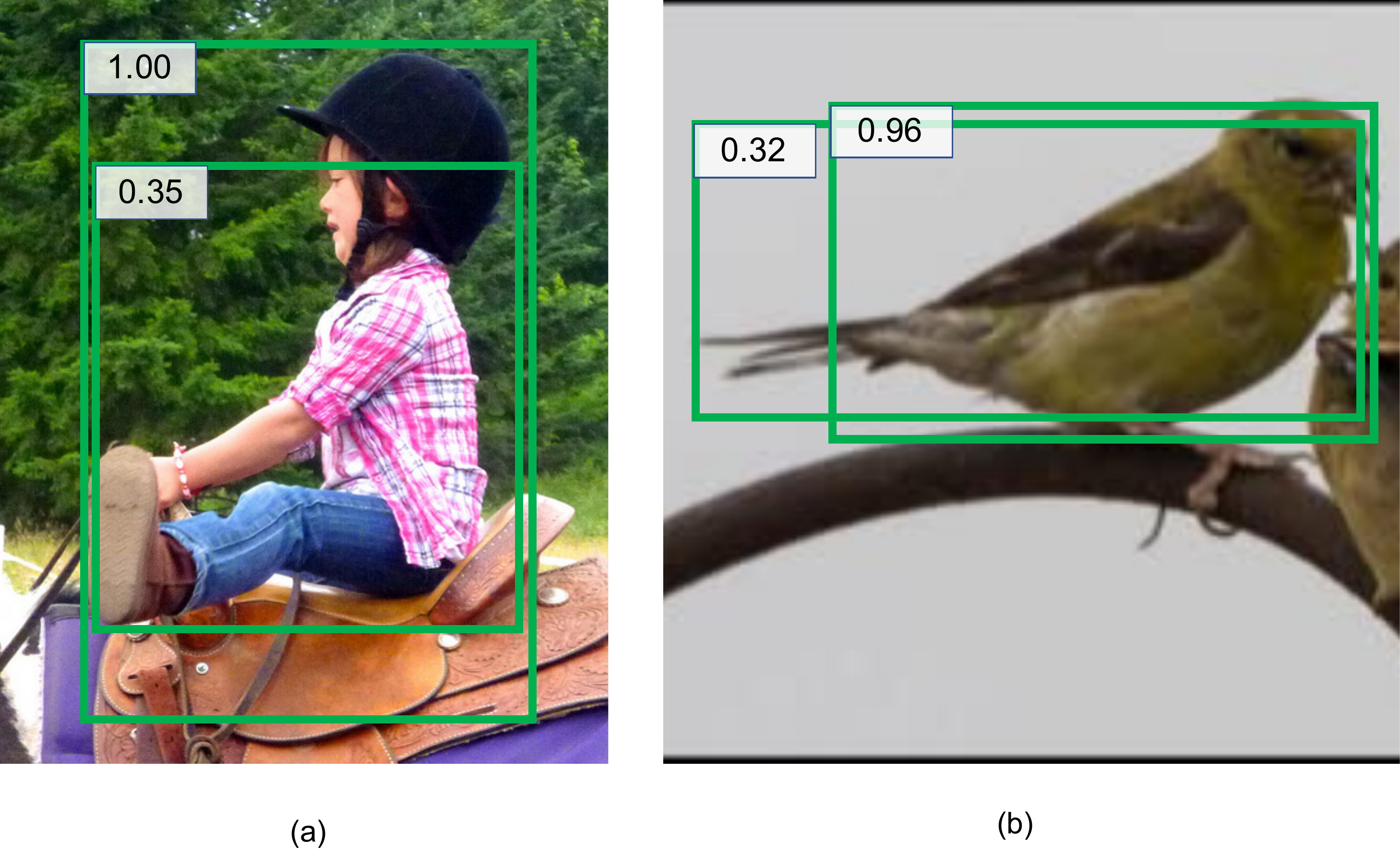}
    \caption{Illustration of failure cases of VGG-16 Faster R-CNN on MS-COCO. (a) both candidate boxes are inaccurate in a certain coordinate.  (b) the left boundary of the bounding box which has the higher classification score is inaccurate. (\textit{better viewed in color})
}
    \label{fig:intro}
    \end{center}
\end{figure}

However, the traditional bounding box regression loss (\ie, the smooth L1 loss~\cite{rcnn}) does not take such the ambiguities of the ground truth bounding boxes into account.
Besides, bounding box regression is assumed to be accurate when the classification score is high, which is not always the case, illustrated in Figure~\ref{fig:intro}.

To address these problems, we propose a novel bounding box regression loss, namely \ourloss, for learning bounding box regression and localization \modelvar\ at the same time. Specifically,  to capture the \modelvars\ of bounding box prediction, we first model the bounding box prediction and ground-truth bounding box as Gaussian distribution and Dirac delta function respectively. Then the new bounding box regression loss is defined as the KL divergence of the predicted distribution and ground-truth distribution. Learning with \ourloss\ has three benefits: (1) The \datasetvars\ in a dataset can be successfully captured. The bounding box regressor gets smaller loss from ambiguous bounding boxes. (2) The learned variance is useful during post-processing. We propose \ournms\ (variance voting) to vote the location of a candidate box using its neighbors' locations weighted by the predicted variances during non-maximum suppression (NMS). (3) The learned probability distribution is interpretable. Since it reflects the level of \modelvar\ of the bounding box prediction, it can potentially be helpful in down-stream applications like self-driving cars and robotics~\cite{djuric2018motion,gualtieri2018learning,he2017vehicle}.

To demonstrate the generality of \ourloss\ and \ournms, we evaluate various CNN-based object detectors on both PASCAL VOC 2007 and MS-COCO including VGG-CNN-M-1024, VGG-16, ResNet-50-FPN, and Mask R-CNN. Our experiments suggest that our approach offers better object localization accuracy for CNN-based object detectors. For VGG-16 Faster R-CNN on MS-COCO, we improve the AP from 23.6\% to \textbf{29.1\%}, with only 2ms increased inference latency on the GPU (GTX 1080 Ti). Furthermore, we apply this pipeline to ResNet-50-FPN Mask R-CNN and improve the AP and AP$^{90}$ by \textbf{1.8\%} and \textbf{6.2\%} respectively, which outperforms the previous state-of-the-art bounding box refinement algorithm~\cite{iounet}.

\section{Related Work}%%%%%%%%%%%%%%%%%%%%%%%%%%%%%%%%%%%%%%%%%
\paragraph{Two-stage Detectors:} Although one-stage detection algorithms~\cite{ssd,yolo,cornernet,fsaf} are efficient, state-of-the-art object detectors are based on two-stage, proposal-driven mechanism~\cite{faster,rfcn,dcn,mask,light,cascade}. Two-stage detectors generate cluttered object proposals, which result in a large number of duplicate bounding boxes. However, during the standard NMS procedure, bounding boxes with lower classification scores will be discarded even if their locations are accurate. Our \ournms\ tries to utilize neighboring bounding boxes based on localization confidence for better localization of the selected boxes.

\paragraph{Object Detection Loss:} To better learn object detection, different kind of losses have been proposed. UnitBox~\cite{unitbox} introduced an Intersection over Union (IoU) loss function for bounding box prediction. Focal Loss~\cite{fl} deals with the class imbalance by changing the standard cross entropy loss such that well-classified examples are assigned lower weights. \cite{rao2018learning} optimizes for the mAP via policy gradient for learning globally optimized object detector. \cite{uncertainties} introduces uncertainties for depth estimation. The idea is further extended to the 3D object detection~\cite{feng2018leveraging,feng2018towards}.  \cite{kendall2017multi} proposes to weight multi-task loss for scene understanding by considering the uncertainty of each task. %, 
With \ourloss, our model can adaptively adjust variances for the boundaries of every object during training, which can help to learn more discriminative features.

\paragraph{Non-Maximum Suppression:} NMS has been an essential part of computer vision for many decades. It is widely used in edge detection~\cite{rosenfeld1971edge}, feature point detection~\cite{lowe2004distinctive} and objection detection~\cite{rcnn,fast,faster,rothe2014non}.
Recently, soft NMS and learning NMS~\cite{softnms,nmslearn} are proposed to improve NMS results. Instead of eliminating all lower scored surrounding bounding boxes, soft-NMS~\cite{softnms} decays the detection scores of all other neighbors as a continuous function of their overlap with the higher scored bounding box. Learning NMS~\cite{nmslearn} proposed to learn a new neural network to perform NMS using only boxes and their classification scores. 

\paragraph{Bounding Box Refinement:} MR-CNN~\cite{nmsavg} is first proposed to merge boxes during iterative localization. Relation network~\cite{relationnet} proposes to learn the relation between bounding boxes.  Recently, IoU-Net~\cite{iounet} proposes to learn the IoU between the predicted bounding box and the ground-truth bounding box. IoU-NMS is then applied to the detection boxes, guided by the learned IoU. Different from IoU-Net, we propose to learn the localization variance from a probabilistic perspective. It enables us to learn the variances for the four coordinates of a predicted bounding box separately instead of only IoU.
Our \ournms\ determine the new location of a selected box based on the variances of neighboring bounding boxes learned by \ourloss, which can work together with soft-NMS (Table~\ref{tab:cocovgg} and Table \ref{tab:cocofpn}).

\section{Approach}%%%%%%%%%%%%%%%%%%%%%%%%%%%%%%%%%%%%%%%%%
In this section,  we first introduce our bounding box parameterization. Then we propose \ourloss\ for training detection network with localization confidence. Finally, a new NMS approach is introduced for improving localization accuracy with our confidence estimation.

\subsection{Bounding Box Parameterization}
\begin{figure}[t]
\begin{center}
\includegraphics[width=\linewidth]{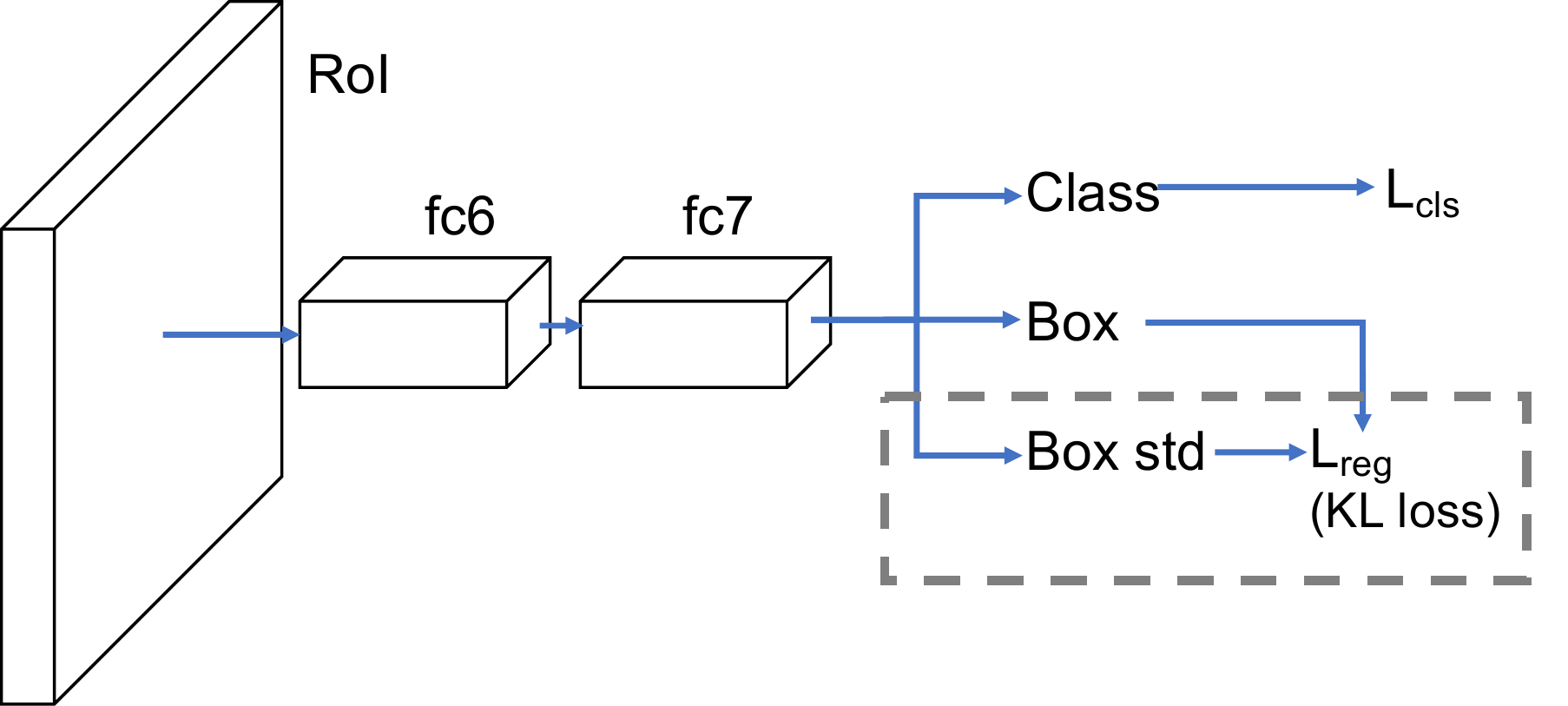}
\caption{Our network architecture for estimating localization confidence. Different from standard fast R-CNN head of a two stage detection network, our network esitmates standard deviations along with bounding box locations, which are taken into account in our regression loss \ourloss}
\label{fig:arch}
\end{center}
\end{figure}

Based on a two-stage object detector Faster R-CNN or Mask R-CNN~\cite{faster,mask} shown in Figure~\ref{fig:arch}, we propose to regress the boundaries of a bounding box separately.
Let $(x_1, y_1, x_2, y_2) \in \mathcal{R}^4$ be the bounding box representation as a 4-dimensional vector, where each dimension is the box boundary location. We adopt the parameterizations of the $(x_1, y_1 ,x_2, y_2)$ coordinates instead of the $(x, y, w, h)$ coordinates used by R-CNN~\cite{rcnn}:
\begin{equation} \label{eq:box_encoding}
\begin{aligned}
& t_{x_1} =\frac{x_1 - x_{1a}}{w_a}, t_{x_2} =\frac{x_2-x_{2a}}{w_a}\\
& t_{y_1} =\frac{y_1 - y_{1a}}{h_a}, t_{y_2} =\frac{y_2-y_{2a}}{h_a}\\
& t_{x_1}^* =\frac{x_1^* - x_{1a}}{w_a}, t_{x_2}^* =\frac{x_2^*-x_{2a}}{w_a}\\
& t_{y_1}^* =\frac{y_1^* - y_{1a}}{h_a}, t_{y_2}^* =\frac{y_2^*-y_{2a}}{h_a}\\
\end{aligned}
\end{equation}
where $t_{x_1}$, $t_{y_1}$, $t_{x_2}$, $t_{y_2}$ are the predicted offsets. $t_{x_1}^*$, $t_{y_1}^*$, $t_{x_2}^*$, $t_{y_2}^*$ are the ground-truth offsets. $x_{1a}$, $x_{2a}$, $y_{1a}$, $y_{2a}$, $w_a$, $h_a$ are from the anchor box. $x_1$, $y_1$, $x_2$, $y_2$ are from the predicted box. 
% Note 
In the following discussions, a bounding box coordinate is denoted as $x$ for simplicity because we can optimize each coordinate independently. 

We aim to estimate the localization confidence along with the location. Formally, our network predicts a probability distribution instead of only bounding box location. Though the distribution could be more complex ones like multivariate Gaussian or a mixture of Gaussians, in this paper we assume the coordinates are independent and use single variate gaussian for simplicity:
\begin{equation}\label{eq:gaussian}
    P_\Theta(x) = \frac{1}{\sqrt{2\pi\sigma^2}} e^{-\frac{(x-x_e)^2}{2\sigma^2}}
\end{equation}
where $\Theta$ is the set of learnable parameters. $x_e$ is the estimated bounding box location. Standard deviation $\sigma$ measures uncertainty of the estimation. When $\sigma \rightarrow 0$, it means our network is extremely confident about estimated bounding box location. It is produced by a fully-connected layer on top of the fast R-CNN head (\verb|fc7|). Figure~\ref{fig:arch} illustrates the fast R-CNN head of our network architecture for object detection. 

The ground-truth bounding box can also be formulated as a Gaussian distribution, with $\sigma \rightarrow 0$, which is a Dirac delta function:
\begin{equation}\label{eq:delta}
    P_D(x) = \delta(x - x_g)
\end{equation}
where $x_g$ is the ground-truth bounding box location.

\subsection{Bounding Box Regression with \ourloss}%-----------------------------------------------

The goal of object localization in our context is to estimate $\hat{\Theta}$ that minimize the KL-Divergence between $P_\Theta(x)$ and $P_D(x)$~\cite{ml} over $N$ samples:
\begin{equation}
 \hat{\Theta} = \argmin_\Theta \frac{1}{N} \sum D_{KL}(P_D(x)||P_\Theta(x))
\end{equation}
We use the KL-Divergence as the loss function $L_{reg}$ for bounding box regression. The classification loss $L_{cls}$ remains the same. For a single sample:
\begin{equation}
\begin{aligned}
   L_{reg} & = D_{KL}(P_D(x)||P_\Theta(x)) \\
    & = \int P_D(x) \log P_D(x) \text{d}x - \int P_D(x) \log P_\Theta(x) \text{d}x\\
    & = \frac{(x_g-x_e)^2}{2\sigma^2} + \frac{\log(\sigma^2)}{2} + \frac{\log(2\pi)}{2} -  H(P_D(x)) 
\end{aligned}
\end{equation}
\begin{figure}[t]
    \begin{center}
    \includegraphics[width=\linewidth]{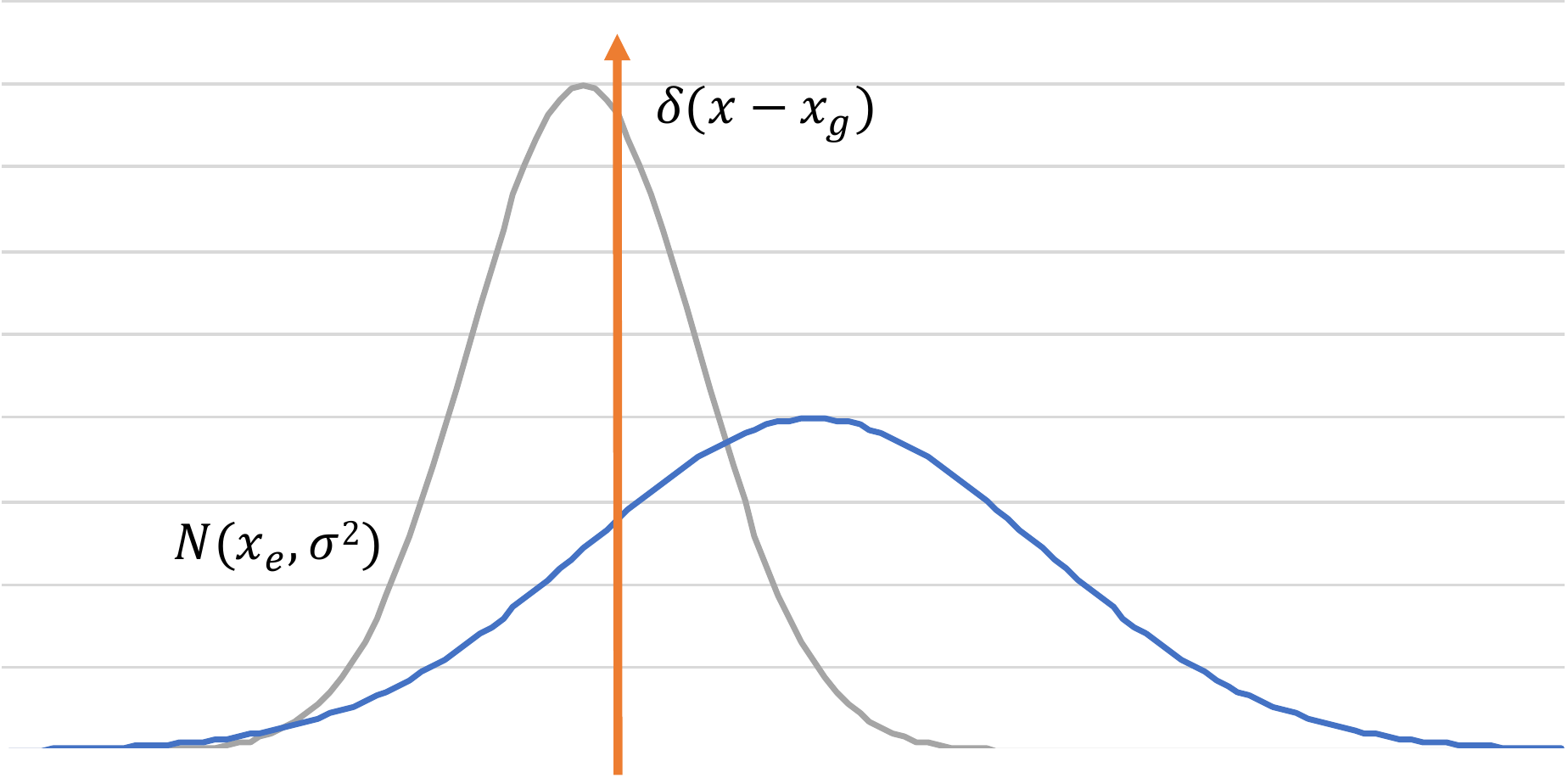}
    \caption{The Gaussian distributions in {\color{blue} blue} and {\color{Gray} gray} are our estimations. The Dirac delta function in {\color{Orange} orange} is the distribution of the ground-truth bounding box. When the location $x_e$ is estimated inaccurately,  we expect the network to be able to predict larger variance $\sigma^2$ so that $L_{reg}$ will be lower (blue)}
    \label{fig:dist}
    \end{center}
\end{figure}
Shown in Figure~\ref{fig:dist}, when the location $x_e$ is estimated inaccurately,  we expect the network to be able to predict larger variance $\sigma^2$ so that $L_{reg}$ will be lower.
$\log(2\pi)/2 $ and $ H(P_D(x)) $ do not depend on the estimated parameters $\Theta$, hence:
\begin{equation}
\begin{aligned}
& L_{reg} \propto \frac{(x_g-x_e)^2}{2\sigma^2} + \frac{1}{2}\log(\sigma^2)
\end{aligned}
\end{equation}
When $\sigma = 1$, \ourloss\ degenerates to the standard Euclidean loss:
\begin{equation}\label{eq:degenerate}
\begin{aligned}
L_{reg} \propto \frac{(x_g-x_e)^2}{2}
\end{aligned}
\end{equation}
The loss is differentiable w.r.t location estimation $x_e$ and localization standard deviation $\sigma$:
\begin{equation}\label{eq:gradient}
\begin{aligned}
\frac{\text{d}}{\text{d}x_e}L_{reg} &= \frac{x_e - x_g}{\sigma^2}\\
\frac{\text{d}}{\text{d}\sigma}L_{reg} &= -\frac{(x_e - x_g)^2}{\sigma^{3}} + \frac{1}{\sigma}
\end{aligned}
\end{equation}
However, since $\sigma$ is in the denominators, the gradient sometimes can explode at the beginning of training. To avoid gradient exploding, our network predicts $\alpha = \log(\sigma^2)$ instead of $\sigma$ in practice:
\begin{equation}\label{eq:l11}
\begin{aligned}
& L_{reg} \propto \frac{e^{-\alpha}}{2}(x_g-x_e)^2 + \frac{1}{2}\alpha\\
\end{aligned}
\end{equation}
We convert $\alpha$ back to $\sigma$ during testing.

For $|x_g-x_e| > 1$, we adopt a loss similar to the smooth $L_1$ loss defined in Fast R-CNN~\cite{fast}:
\begin{equation}\label{eq:l12}
L_{reg} = e^{-\alpha}(|x_g-x_e|-\frac{1}{2}) + \frac{1}{2}\alpha
\end{equation}

We initialize the weights of the FC layer for $\alpha$ prediction with random Gaussian initialization. The standard deviation and mean are set to 0.0001 and 0 respectively, so that \ourloss\ will be similar to the standard smooth L1 loss at the beginning of training. (Equation~\ref{eq:l11} and Equation~\ref{eq:l12}). 

\subsection{\Ournms}\label{sec:ournms}%---------------------------------------------------------------------
\begin{figure*}[t] 
\begin{center}
\includegraphics[width=\linewidth]{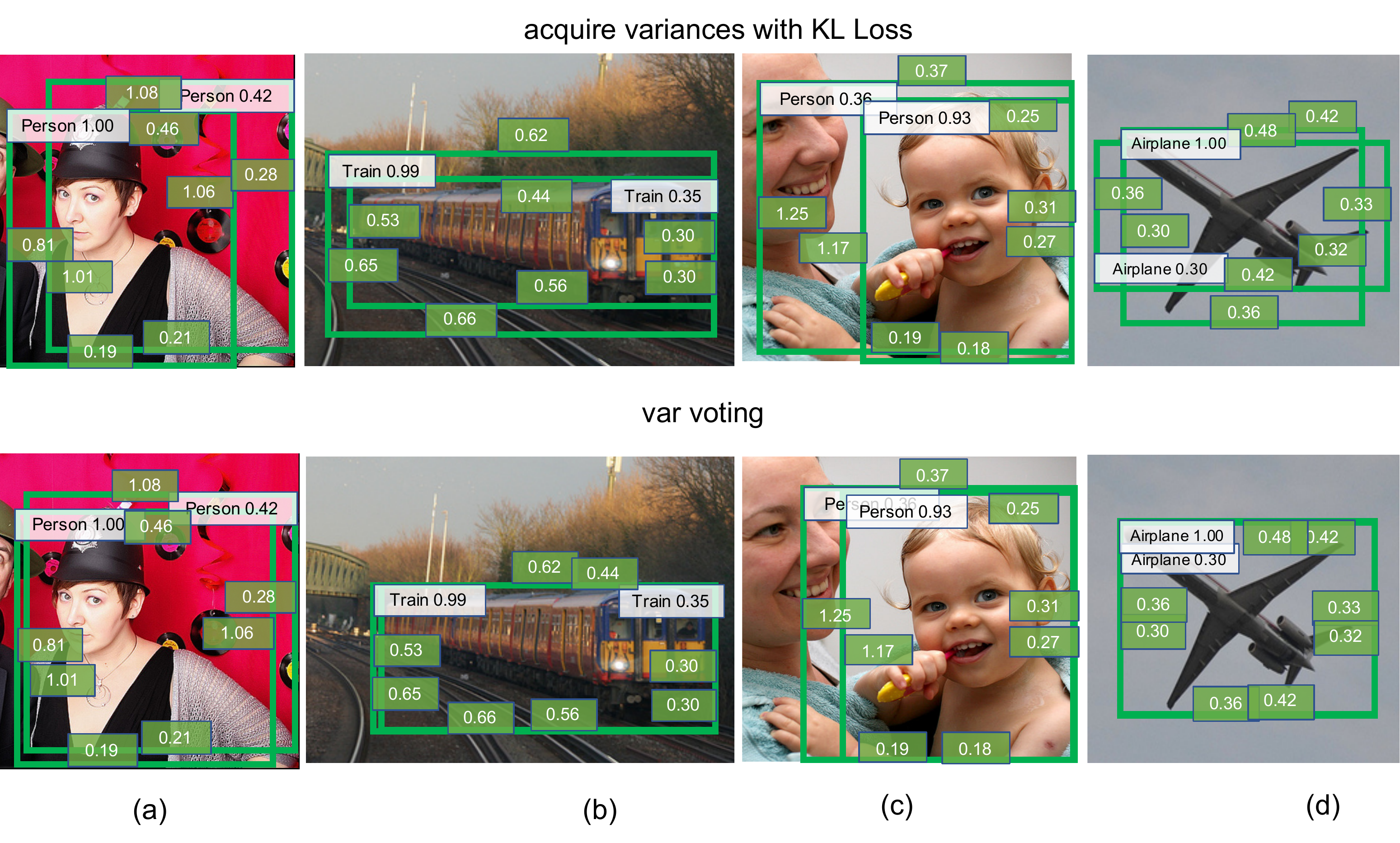}
    \caption{Results of \ournms\ with VGG-16 Faster R-CNN on MS-COCO. The {\color{ForestGreen} green} textbox in the middle of each boundary is the corresponding standard deviation $\sigma$ we predicted (Equation~\ref{eq:gaussian}). Two failure situations corresponding to Figure~\ref{fig:intro} that can be improved by \ournms: (a) When each candidate bounding box is inaccurate in some coordinates (women on the right), our \ournms\ can incorporate their localization confidence and produce better boxes. (b) The bounding box with a higher classification score (train 0.99) actually has lower localization confidence than the bounding box with a lower classification score (train 0.35). After \ournms, the box scored 0.99 moves towards the correct location. (\textit{better viewed in color})} 
    \label{fig:img} 
    \end{center}
 \end{figure*}
 After we obtain the variance of predicted location, it is intuitive to vote candidate bounding box location according to the learned variances of neighboring bounding boxes.
Shown in Algorithm~\ref{alg:nms}, we change NMS with three lines of code:
\begin{algorithm}[H]
\caption {\ournms}{$\mathcal{B}$ is $N\times4$ matrix of initial detection boxes.
$\mathcal{S}$ contains corresponding detection scores.
$\mathcal{C}$ is $N\times4$ matrix of corresponding variances. $\mathcal{D}$ is the final set of detections.
$\sigma_t$ is a tunable parameter of \ournms. The lines in {\color{blue}blue} and in {\color{ForestGreen}green} are soft-NMS and \ournms\ respectively.}\label{alg:nms}
\begin{algorithmic}
\State $\mathcal{B} = \{b_1, ..,b_N\} $, $\mathcal{S} = \{s_1, ..,s_N\}$, $\mathcal{C} = \{\sigma^2_1, ..,\sigma^2_N\}$ %, $\sigma_t$
\State$\mathcal{D} \leftarrow \{\}$
\State$\mathcal{T} \leftarrow \mathcal{B}$
\While {$\mathcal{T} \ne$ empty}
\State$m \leftarrow$ argmax $\mathcal{S} $
% \State$\mathcal{M} \leftarrow$ $b_m$ 
\State $ \mathcal{T} \leftarrow \mathcal{T} - b_m$ %\mathcal{M}
\State\textcolor{blue}{$ \mathcal{S} \leftarrow \mathcal{S} f(IoU(b_m, T)) $} \Comment{soft-NMS}
\textcolor{ForestGreen}{
\State$idx  \leftarrow IoU(b_m, B) > 0$} \Comment{\ournms}%\mathcal{M}
\textcolor{ForestGreen}{
\State$p  \leftarrow exp(-(1-IoU(b_m, \mathcal{B}[idx]))^2/\sigma_t)$}
\textcolor{ForestGreen}{
\State $b_m \leftarrow p(\mathcal{B}[idx]/\mathcal{C}[idx])/p(1/\mathcal{C}[idx])$} %\mathcal{M}
\State$\mathcal{D} \leftarrow \mathcal{D} \bigcup b_m$%\mathcal{M }
\EndWhile
\State\textbf{return} {$\mathcal{D}, \mathcal{S}$}
\end{algorithmic}
\end{algorithm}
 We vote the location of selected boxes within the loop of standard NMS or soft-NMS~\cite{softnms}. After selecting the detection with maximum score $b$,
 $\{x_1, y_1, x_2, y_2, s, \sigma_{x_1}, \sigma_{y_1}, \sigma_{x_2}, \sigma_{y_2}\}$, its new location is computed according to itself and its neighboring bounding boxes. %\mathcal{M}
Inspired by soft-NMS, we assign higher weights for boxes that are closer and have lower uncertainties. Formally, let $x$ be a coordinate (\eg, $x_1$) and $x_i$ be the coordinate of $i$th box. The new coordinate is computed as follow:
\begin{equation}
    \begin{aligned}
p_i &= e^{-(1-IoU(b_i,b))^2/\sigma_t}\\ %\mathcal{M}
x &= \frac{\sum_i p_ix_i / \sigma_{x,i}^2}{\sum_i p_i/\sigma_{x,i}^2}\\
& \text{subject\ to } IoU(b_i, b) > 0 %\mathcal{M}
    \end{aligned}
\end{equation}
$\sigma_t$ is a tunable parameter of \ournms. Two types of neighboring bounding boxes will get lower weights during voting: (1)
Boxes with high variances. (2) Boxes that have small IoU with the selected box.  Classification score is not involved in the voting, since lower scored boxes may have higher localization confidence. In Figure~\ref{fig:img}, we provide a visual illustration of \ournms. With \ournms, the two situations as mentioned earlier in Figure~\ref{fig:intro} that lead to detection failure can sometimes be avoided.

\section{Experiments}%%%%%%%%%%%%%%%%%%%%%%%%%%%%%%%%%%%%%%%%%
To demonstrate our method for accurate object detection, we use two datasets: MS-COCO~\cite{coco} and PASCAL VOC 2007~\cite{pascal-voc-2007}.
We use four GPUs for our experiments. The training schedule and batch size are adjusted according to the linear scaling rule~\cite{hour}. 
For VGG-CNN-M-1024 and VGG-16 Net~\cite{vgg}, our implementation is based on Caffe~\cite{caffe}.
For ResNet-50 FPN~\cite{resnet,fpn} and Mask R-CNN~\cite{mask}, our implementation is based on Detectron~\cite{detectron}.
For VGG-16~\cite{vgg} Faster R-CNN, following \verb|py-faster-rcnn|\footnote{github.com/rbgirshick/py-faster-rcnn}, we train on \verb|train2014| and test on \verb|val2014|. For other object detection networks, we train and test on the newly defined \verb|train2017| and \verb|val2017| respectively. We set $\sigma_t$ to 0.02. Unless specified, all hyper-parameters are set to default\footnote{github.com/facebookresearch/Detectron}.

\begin{table*}[t]
    \begin{center}
    \begin{tabular}{c|c|c|c|c|c|c|c|c|c|c|c}
    \hline\hline
    \ourloss & soft-NMS & \ournms  & AP & AP$^{50}$ & AP$^{75}$ & AP$^S$ & AP$^M$ & AP$^L$ & AR$^{1}$ & AR$^{10}$ & AR$^{100}$  \\ \hline\hline
 & &            &23.6 & 44.6& 22.8 &6.7 &25.9 &36.3 &23.3 &33.6 &  34.3  \\ \hline
&\checkmark & & 24.8 & 45.6 & 24.6 & 7.6 & 27.2 & 37.6 & 23.4 & 39.2 & 42.2 \\ \hline
    \checkmark & & & 26.4 & 47.9 & 26.4 & 7.4 & 29.3 & 41.2 & 25.2 & 36.1 & 36.9 \\ \hline
\checkmark & &\checkmark & 27.8 & 48.0 & 28.9 & 8.1 & 31.4 & 42.6 & 26.2 & 37.5 & 38.3 \\ \hline
\checkmark &\checkmark & & 27.8 & 49.0 & 28.5 & 8.4 & 30.9 & 42.7 & 25.3 & 41.7 & 44.9 \\ \hline
\checkmark &\checkmark &\checkmark &\textbf{29.1} &\textbf{49.1} &\textbf{30.4} &\textbf{8.7} &\textbf{32.7} &\textbf{44.3} &\textbf{26.2} &\textbf{42.5}&\textbf{45.5} \\ \hline
    \end{tabular}
    
\caption{The contribution of each element in our detection pipeline on MS-COCO. The baseline model is VGG-16 Faster R-CNN}
\label{tab:cocovgg}
\end{center}
\end{table*}

\begin{table}[t]
\begin{center}
\begin{tabular}{c|c}
\hline\hline
method & latency (ms) \\ \hline
baseline & 99 \\ \hline
ours & 101 \\ \hline
\end{tabular}
\end{center}
\caption{Inference time comparison on MS-COCO with VGG-16 Faster R-CNN on a GTX 1080 Ti GPU, CUDA 8~\cite{cuda} and CUDNN 6~\cite{cudnn}}
\label{tab:cocotime}
\end{table}
\begin{table}[t]
\begin{center}
    \begin{tabular}{l|c|c|l}
\hline\hline
fast R-CNN head & backbone & \ourloss & \multicolumn{1}{c}{AP} \\ \hline
\multirow{2}{*}{2mlp head} & \multirow{2}{*}{FPN} &  & 37.9 \\ %\cline{3-4} 
 &  & \checkmark & 38.5$^{+0.6}$ \\ \hline
 \multirow{2}{*}{2mlp head  + mask} & \multirow{2}{*}{FPN} &  & 38.6 \\ %\cline{3-4} 
 &  & \checkmark & 39.5$^{+\textbf{0.9}}$ \\ \hline
\multirow{2}{*}{conv5 head} & \multirow{2}{*}{RPN} &  & 36.5 \\ %\cline{3-4} 
 &  & \checkmark & 38.0$^{+\textbf{1.5}}$ \\ \hline
\end{tabular}
\caption{Comparison of different fast R-CNN heads. The model is ResNet-50 Faster R-CNN}
\label{tab:c5}
\end{center}
\end{table}

\subsection{Ablation Study}
We evaluate the contribution of each element in our detection pipeline: 
% \verb|x1, y1, x2, y2| coordinates,
\ourloss, soft-NMS and \ournms\ with VGG-16 Faster R-CNN. The detailed results are shown in Table~\ref{tab:cocovgg}.

\paragraph{\ourloss:} Surprisingly, simply training with \ourloss\ greatly improves the AP by \textbf{2.8\%}, which is also observed on ResNet-50 Faster R-CNN and Mask R-CNN (\textbf{1.5\%} and \textbf{0.9\%} improvement respectively, shown in Table~\ref{tab:c5} and Table~\ref{tab:mask}). First, by learning to predict high variances for samples with high uncertainties during training, the network can learn more from useful samples. Second, the gradient for localization can be adaptively controlled by the network during training (Equation~\ref{eq:gradient}), which encourages the network to learn more accurate object localization. Third, \ourloss\ incorporates learning localization confidence which can potentially help the network to learn more discriminative features. 

The learned variances through our \ourloss\ are interpretable. Our network will output higher variances for challenging object boundaries, which can be useful in vision applications like self-driving cars and robotics. The first row of Figure~\ref{fig:img} shows some qualitative examples of the standard deviation learned through our \ourloss.

\paragraph{Soft-NMS:} As expected, soft-NMS performs consistently on both baseline and our network trained with \ourloss. It improves the AP by 1.2\% and 1.4\% on the baseline and our network respectively, shown in Table~\ref{tab:cocovgg}.

\paragraph{\Ournms:} Finally, with \ournms, the AP is further improved to \textbf{29.1\%}. We made the observation that improvement mainly comes from the more accurate localization. Notice that the AP$^{50}$ is only improved by 0.1\%. However, AP$^{75}$, AP$^{M}$ and AP$^{L}$ are improved by 1.8\%, 1.8\%, and 1.6\% respectively, shown in Table~\ref{tab:cocovgg}.  This indicates that classification confidence is not always associated with localization confidence. Therefore, learning localization confidence apart from classification confidence is important for more accurate object localization. 

We also found that \ournms\ and soft-NMS can work well together. Applying \ournms\ with the standard NMS improves the AP by 1.4\%. Applying \ournms\ after soft-NMS still can improve the AP by 1.3\%. We argue that soft-NMS is good at scoring candidate bounding boxes which improve overall performance, whereas \ournms\ is good at refining those selected bounding boxes for more accurate object localization. The second row of Figure~\ref{fig:img} shows some qualitative examples of our \ournms.
\begin{figure}[t]
    \begin{center}
    \includegraphics[width=\linewidth]{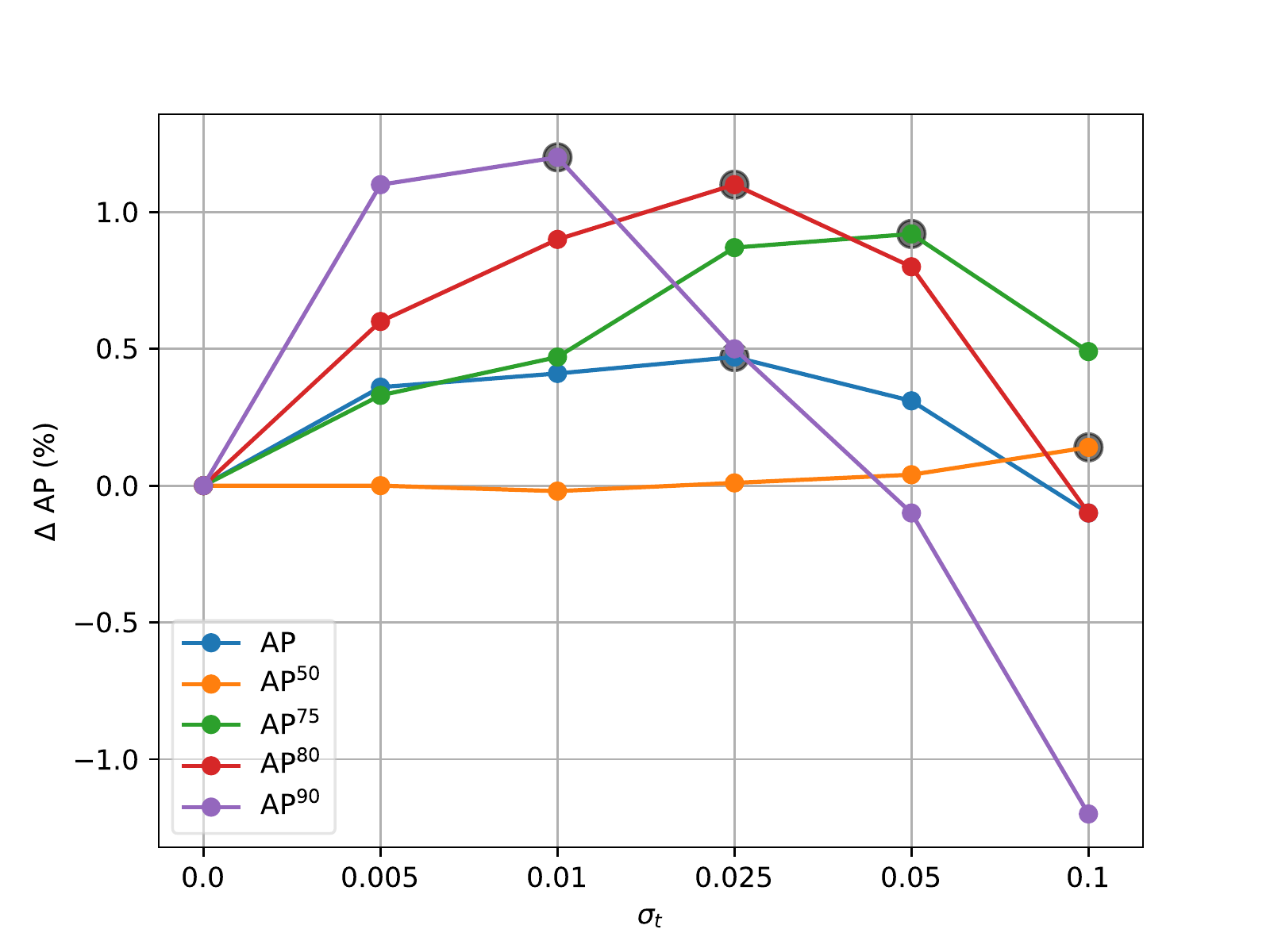}
    \caption{Varying $\sigma_t$ for \ournms\ with ResNet-50 Faster R-CNN. (\textit{better viewed in color})}
    \label{fig:sigma}
    \end{center}
\end{figure}

Shown in Figure~\ref{fig:sigma}, we test the sensitivity of the tunable parameter $\sigma_t$ for \ournms. When $\sigma_t=0$, \ournms\ is not activated. We observe that the AP$^{75}$, AP$^{80}$ and AP$^{90}$ can be significantly affected by $\sigma_t$, while AP$^{50}$ is less sensitive to $\sigma_t$. Acceptable values of $\sigma_t$ varies from around $0.005 \sim 0.05$. We use $\sigma_t=0.02$ in all experiments. 

\paragraph{Inference Latency:}
We also evaluate the inference time of our improved VGG-16 Faster R-CNN on a single GTX 1080 Ti GPU with CUDA 8 and CUDNN 6, as it is crucial for resource-limited applications~\cite{shufflenet,addressnet,cp,amc,lightweight}.
Shown in Table~\ref{tab:cocotime}, our approach only increases \textbf{2ms} latency on GPU. Different from IoUNet~\cite{iounet} which uses \verb|2mlp| head for IoU prediction, our approach only requires a $4096\times324$ fully-connected layer for the localization confidence prediction. 
% As for NMS processing, our implementation based on single thread Python is slow. However, it can be significantly optimized by implementing in CUDA or C++~\cite{detectron,cuda,cudnn}.

\paragraph{RoI Box Head:}
We test the effectiveness of \ourloss\ with different RoI box heads on a deeper backbone: ResNet-50. \verb|res5/conv5| head consists of 9 convolutional layers which can be applied to each RoI as fast R-CNN head. \verb|2mlp| head consists of two fully connected layers.  \verb|res5| head can learn more complex representation than the commonly used \verb|2mlp| head.  Shown in Table~\ref{tab:c5}, \ourloss\ can improve the AP by \textbf{0.9\%} with mask. \ourloss\ can further improve the AP by \textbf{1.5\%} with \verb|conv5| head. We hypothesize that the localization variance is much more challenging to learn than localization, therefore \ourloss\ can benefit more from the expressiveness of \verb|conv5| head. Since \verb|conv5| head is not commonly used in recent state-of-the-art detectors, we still adopt the \verb|2mlp| head in the following experiments.

\begin{table*}[t]
\begin{center}
\begin{tabular}{l|l|lllll}
\hline\hline
 & \multicolumn{1}{c|}{AP} & \multicolumn{1}{c}{AP$^{50}$} & \multicolumn{1}{c}{AP$^{60}$} & \multicolumn{1}{c}{AP$^{70}$} & \multicolumn{1}{c}{AP$^{80}$} & \multicolumn{1}{c}{AP$^{90}$} \\ \hline
baseline~\cite{detectron} & 38.6 & \textbf{59.8} & 55.3 & 47.7 & 34.4 & 11.3 \\
MR-CNN~\cite{nmsavg} & 38.9 & \textbf{59.8} & 55.5 & 48.1 &  34.8$^{+0.4}$ & 11.9$^{+0.6}$ \\
soft-NMS~\cite{softnms} & 39.3 & 59.7 & \textbf{55.6} & \textbf{48.9} &35.9$^{+1.5}$ &12.0$^{+0.7}$ \\
IoU-NMS+Refine~\cite{iounet} & 39.2 & 57.9 & 53.6 & 47.4 & 36.5$^{+2.1}$ & 16.4$^{+5.1}$ \\ \hline
\ourloss & 39.5$^{+0.9}$ & 58.9 & 54.4 & 47.6 & 36.0$^{+1.6}$ & 15.8$^{+4.5}$ \\
\ourloss+\ournms & 39.9$^{+1.3}$ & 58.9 & 54.4 & 47.7 & 36.4$^{+2.0}$ & 17.0$^{+5.7}$ \\
\ourloss+\ournms+soft-NMS & \textbf{40.4}$^{+1.8}$ & 58.7 & 54.6 & 48.5 & \textbf{37.5}$^{+3.3}$ & \textbf{17.5}$^{+6.2}$ \\ \hline
\end{tabular}
\caption{Comparisons of different methods for accurate object detection on MS-COCO. The baseline model is ResNet-50-FPN Mask R-CNN. We improve the baseline by $\approx$ 2\% in AP}
\label{tab:mask}
\end{center}
\end{table*}
\subsection{Accurate Object Detection}
Table~\ref{tab:mask} summarizes the performance of different methods for accurate object detection on ResNet-50-FPN Mask R-CNN. With \ourloss, the network can learn to adjust the gradient for ambiguous bounding boxes during training. As a result, Mask R-CNN trained with \ourloss\ performs significantly better than the baseline for high overlap metrics like AP$^{90}$. \Ournms\ improves the localization results by voting the location according to the localization confidences of neighboring bounding boxes. AP$^{80}$ and AP$^{90}$ are further improved by 0.4\% and 1.2\% respectively. \Ournms\ is also compatible with soft-NMS. \Ournms\ combined with soft-NMS improves the AP$^{90}$ and the overall AP of the final model by \textbf{6.2\%} and \textbf{1.8\%} respectively. Compared with IoUNet~\cite{iounet}:  (1)  our variance and localization are learned together with \ourloss, which improves the performance. (2) \ourloss\ does not require a separate 2mlp head for learning localization confidence, which introduces nearly no additional computation. (3) \ournms\ does not require iterative refinement, which is much faster.

\begin{table}[t]
    \begin{center}
\begin{tabular}{c|l|c}
\hline\hline
backbone & \multicolumn{1}{c|}{method} & mAP \\ \hline
\multirow{4}{*}{} & baseline & 60.4 \\ \cline{2-3} 
VGG-CNN- & \ourloss & 62.0 \\ %\cline{2-3} 
M-1024 & \ourloss+\ournms & 62.8 \\ %\cline{2-3} 
 & \ourloss+\ournms+soft-NMS & \textbf{63.6} \\ \hline
\multirow{7}{*}{VGG-16} 
& baseline & 68.7 \\ %\cline{2-3} 
  & QUBO (tabu)~\cite{qubo} & 60.6 \\ %\cline{2-3} 
  & QUBO (greedy)~\cite{qubo} & 61.9 \\ %\cline{2-3} 
 & \softnms & 70.1 \\ \cline{2-3} 
 & \ourloss & 69.7 \\ %\cline{2-3}  
 & \ourloss+\ournms & 70.2 \\ %\cline{2-3} 
 & \ourloss+\ournms+soft-NMS & \textbf{71.6} \\ \hline
\end{tabular}
\caption{Comparisons of different approaches on PASCAL VOC 2007 with Faster R-CNN.}
\label{tab:vocvgg}
\end{center}
\end{table}

\begin{table*}[t]
\begin{center}
    \begin{tabular}{c|l|c|ccccc}\hline\hline
type & \multicolumn{1}{c|}{method} & AP & AP$^{50}$ & AP$^{75}$ & AP$^S$ & AP$^M$& AP$^L$  \\ \hline
\multirow{7}{*}{fast R-CNN}  & baseline (1x schedule)~\cite{detectron}  &36.4 &\textbf{58.4} &39.3 &\textbf{20.3} &39.8 &48.1\\
 & baseline (2x schedule)~\cite{detectron} &36.8 &\textbf{58.4} &39.5 &19.8 &39.5 &49.5 \\
  & IoU-NMS~\cite{iounet}  & 37.3  & 56.0 &-  &-  &-  &- \\ 
& soft-NMS~\cite{softnms} &37.4 &58.2 &41.0 &\textbf{20.3} &40.2 &50.1\\ \cline{2-8} 
 & \ourloss &37.2 &57.2 &39.9 &19.8 &39.7 &50.1\\ 
 & \ourloss+\ournms  &37.5 &56.5 &40.1 &19.4 &40.2 &51.6\\ 
& \ourloss+\ournms+soft-NMS  &\textbf{38.0} &56.4 &\textbf{41.2} &19.8 &\textbf{40.6} &\textbf{52.3}\\ \hline\hline
\multirow{9}{*}{Faster R-CNN}  &baseline (1x schedule)~\cite{detectron} &36.7 &58.4 &39.6 &21.1 &39.8 &48.1\\
 & IoU-Net~\cite{iounet}  & 37.0 &58.3  & - & -&-&- \\ 
     & IoU-Net+IoU-NMS~\cite{iounet}        & 37.6& 56.2 &-&-&- &- \\ 
    & baseline (2x schedule)~\cite{detectron}    &37.9 &59.2 &41.1 &21.5 &41.1 &49.9 \\
 & IoU-Net+IoU-NMS+Refine~\cite{iounet}   &38.1  & 56.3&-&-&-&-  \\ 
% & 160k & voting  &38.2 &59.3 &41.3 &21.7 &41.3 &50.7 \\ \cline{3-9} 
  & soft-NMS\cite{softnms}       &38.6 &\textbf{59.3} &42.4 &21.9 &\textbf{41.9} &50.7 \\ \cline{2-8} 
 & \ourloss  & 38.5 & 57.8& 41.2 & 20.9 & 41.2 & 51.5\\  
 & \ourloss +\ournms  & 38.8 & 57.8& 41.6 & 21.0 & 41.5 & 52.0\\ 
 & \ourloss +\ournms+soft-NMS  & \textbf{39.2} & 57.6& \textbf{42.5} & 21.2 & 41.8 & \textbf{52.5}\\  \hline

\end{tabular}
\caption{Performance comparison with FPN ResNet-50 on MS-COCO}
\label{tab:cocofpn}
\end{center}
\end{table*}

We further evaluate our approach on the feature pyramid network (ResNet-50 FPN)~\cite{fpn,resnet}, shown in Table~\ref{tab:cocofpn}. 

For fast R-CNN version, training with \ourloss\ increases the baseline by 0.4\%. After applying \ournms\ along with soft-NMS, our model achieves \textbf{38.0\%} in AP, which outperforms both IoU-NMS and soft-NMS baselines. Training end-to-end with \ourloss\ can help the network learn more discriminative features, which improves the baseline AP by \textbf{0.6\%}. The final model achieves 39.2\% in AP, which improves the baseline by 1.3\%.

\subsection{Experiments on PASCAL VOC 2007}%-----------------------------------------------

Even though our approach is designed for large scale object detection, it could also generalize well on small datasets. We perform experiments with Faster R-CNN~\cite{faster} on PASCAL VOC 2007, which consists of about 5k \verb|voc_2007_trainval| images and 5k \verb|voc_2007_test| images over 20 object categories. Backbone networks: VGG-CNN-M-1024 and VGG-16 Net~\cite{vgg} are tested. 

Shown in Table~\ref{tab:vocvgg}, we compare our approach with soft-NMS and quadratic unconstrained binary optimization (QUBO~\cite{qubo}). 
For QUBO, we test both greedy and classical tabu solver (we manually tuned the penalty term for both solvers to get better performance). We observe that it is much worse than the standard NMS, though it was reported to be better for pedestrian detection. We hypothesize that QUBO is better at pedestrian detection since there are more occluded bounding boxes~\cite{crowdhuman}.
For VGG-CNN-M-1024, training with \ournms\ improves the mAP by \textbf{1.6\%}. \ournms\ further improves the mAP by \textbf{0.8\%}.
For VGG-16, our approach improves the mAP by \textbf{2.9\%}, combined with soft-NMS.
We notice that \ournms\ could still improve performance even after soft-NMS is applied to the initial detection boxes. This observation is consistent with our experiments on MS-COCO (Table~\ref{tab:cocovgg}).

\section{Conclusion}%%%%%%%%%%%%%%%%%%%%%%%%%%%%%%%%%%%%%%%%%
To conclude, the uncertainties in large-scale object detection datasets can hinder the performance of state-of-the-art object detectors. Classification confidence is not always strongly related to localization confidence. In this paper, a novel bounding box regression loss with uncertainty is proposed for learning more accurate object localization. By training with \ourloss, the network learns to predict localization variance for each coordinate. The resulting variances empower \ournms, which can refine the selected bounding boxes via voting. Compelling results are demonstrated for VGG-16 Faster R-CNN, ResNet-50 FPN and Mask R-CNN on both MS-COCO and PASCAL VOC 2007. 

\section*{Acknowledgement}%%%%%%%%%%%%%%%%%%%%%%%%%%%%%%%%%%%%%%%%%
This research was partially supported by National Key R\&D Program of China (No.~2017YFA0700800). 

We would love to express our appreciation to Prof.~Kris Kitani and Dr.~Jian Sun for the useful discussions during this research. \nocite{softernms,pts} 

{\small
\bibliographystyle{ieee}
\bibliography{egbib}
}

\end{document}